\pdfoutput=1

\documentclass[11pt]{article}

\usepackage[]{EMNLP2022}

\usepackage{times}
\usepackage{latexsym}
\usepackage{graphicx}
\usepackage{algorithm}
\usepackage{algorithmic}

\usepackage[T1]{fontenc}

\usepackage[utf8]{inputenc}

\usepackage{microtype}
\usepackage{enumitem}
\usepackage{inconsolata}

\title{Interpretable Detection of Out-of-Context Misinformation with Neural-Symbolic-Enhanced Large Multimodal Model}

 \author{Yizhou Zhang,~~ Loc Trinh,~~ Defu Cao,~~ Zijun Cui,~~ Yan Liu  \\
          University of Southern California\\
          \texttt{\{zhangyiz,~loctrinh, defucao, zijuncui, yanliu.cs\}@usc.edu}}

\begin{document}
\maketitle
\begin{abstract}
Recent years have witnessed the sustained evolution of misinformation that aims at manipulating public opinions. Unlike traditional rumors or fake news editors who mainly rely on generated and/or counterfeited images, text and videos, current misinformation creators now more tend to use out-of-context multimedia contents (e.g. mismatched images and captions) to deceive the public and fake news detection systems. This new type of misinformation increases the difficulty of not only detection but also clarification, because every individual modality is close enough to true information. To address this challenge, in this paper we explore how to achieve interpretable cross-modal de-contextualization detection that simultaneously identifies the mismatched pairs and the cross-modal contradictions, which is helpful for fact-check websites to document clarifications. The proposed model first symbolically disassembles the text-modality information to a set of fact queries based on the Abstract Meaning Representation of the caption and then forwards the query-image pairs into a pre-trained large vision-language model select the ``evidences" that are helpful for us to detect misinformation. Extensive experiments indicate that the proposed methodology can provide us with much more interpretable predictions while maintaining the accuracy same as the state-of-the-art model on this task.
\end{abstract}

\section{Introduction}
Increasing exploits of multimedia misinformation (e.g., news with edited images and/or machine-generated text) are threatening the credibility and reliability of online information \cite{shu2017fake,kumar2018false,sharma2019combating,sharma2020coronavirus}. Misinformation campaigns create and spread fake news and/or rumors with specific topics and narratives to manipulate public opinions \cite{sharma2020identifying,zhang2021vigdet} in different areas, e.g., healthcare (COVID-19 pandemic and vaccines)\cite{sharma2021covid, zhang2022counterfactual} and politics (elections) \cite{sharma2022characterizing}. To address the challenge, researchers have conducted attempts from different directions, such as detecting the trace of image editing tools (e.g., Photoshop for image and DeepFake for video) \cite{Trinh_2021_WACV} and identifying the linguistic cues of text generated by machine or written by misinformation campaigns \cite{qian2018neural,khattar2019mvae}.

\begin{figure*}
    \centering
    \includegraphics[width=6.3in]{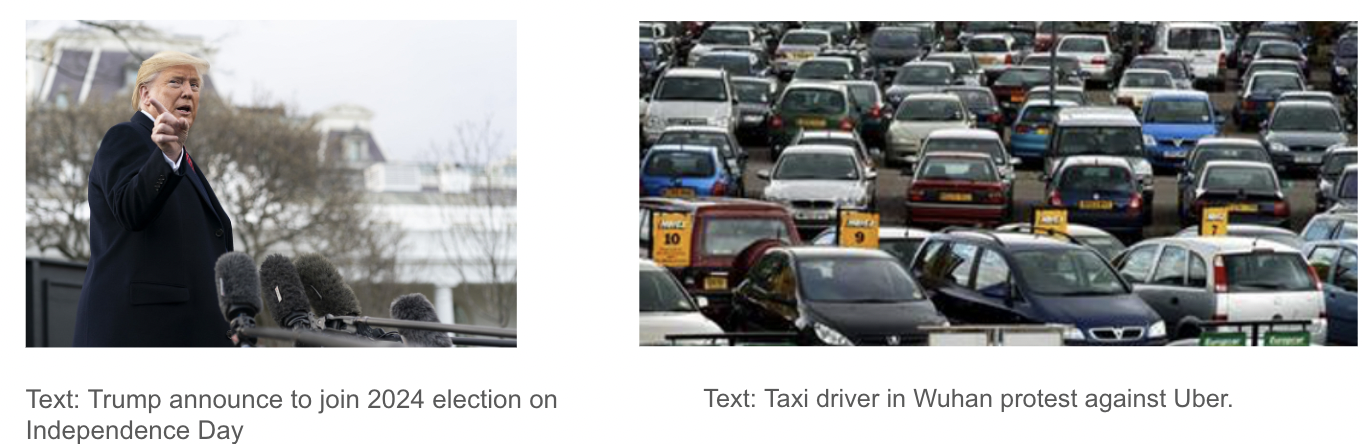}
    \caption{Examples of mismatched text-image pairs. The left pair is mismatched because the image is obviously taken in winter (the clothes and background), rather than Independence Day. And the right pair is mismatched because the cars are with yellow license plates, but in China the taxi uses blue license plates.}
    \label{fig:my_label}
\end{figure*}

However, as researchers are enriching our toolbox, the misinformation creators are also evolving. Text-image de-contextualization \cite{jaiswal2017multimedia} is one of their tools to evade being detected. The misinformation campaigns can recruit experienced human writers to imitate the wording of true news and use unedited but mismatched images. In this way, they can manipulate people's ideas by using fake news with both real text and real images, which greatly confuses the machine learning models based on traces of image editing tools and/or auto-generated text. This challenge is even worse as the development of large pre-trained multi-modal models, which enables misinformation campaigns to automatically retrieve the most deceptive image from a million-scale (or even billion-scale) image database \cite{luo2021newsclippings}. Moreover, our experiments (See Table \ref{tab:comparison} in Sec. 4) indicate that such auto-retrieved deceptive images can easily confuse the traditional multi-modal misinformation detector. To address this challenge, some researchers propose to apply the representation-learning-based multi-modal model to detect the inconsistency between the mismatched text-image pairs \cite{luo2021newsclippings,huang2022text}. More specifically, such methods focus on learning a unified latent representation space where the consistency of a text-image pair can be acquired by computing the distance between text embedding and image embedding. 

Nevertheless, the latent space usually lacks interpretability. In other words, we have no idea about the physical meaning of each dimension in the latent space. As a result, even though the model gives the prediction of consistency or inconsistency, the social media platforms and/or fact-checking websites can not document human-understandable clarifications to justify the model prediction and the further actions adopted based on the prediction, e.g. officially alert the users of the suspicious tweets or posts. Consequently, to stop or resist the misinformation spreading, the platforms still need to recruit human verifiers to collect the evidence to justify further actions \cite{sharma2022construction}. Such a process is time and resource-consuming and may lead to missing the best time point of action.

A commonly applied interpretable multi-modal learning paradigm is semantic-graph-based neural-symbolic learning \cite{yi2018neural,zhu2022generalization}. Its central idea is to represent the inputs from different modalities as the same modality: semantic graph (i.e. scene graph for vision and abstract-meaning-representation graph for text). Then under this shared modality, a graph-learning model or a symbolic-based model can acquire explainable answers by doing explicit reasoning on the extracted graphs. However, this paradigm is confronted of serious challenges when being applied for cross-modal misinformation detection. First, the process that converts vision inputs to graphs may miss some details, such as background. However, with misinformation detection, the key points might be hidden in details, as shown in Figure \ref{fig:my_label}. Second, for misinformation detection, we not only need basic concepts like actions and object labels but also complicated concepts like social relations and identity. Generic scene graph parsers may fail to capture such complicated semantics. 

To address the above drawbacks of existing works, in this paper, we propose an interpretable cross-modal de-contextualization detector. The proposed model not only needs to predict whether a given text-image pair is consistent or not, but also needs to output the pieces of  ``evidence'' that can support the prediction. In this way, we can ease the workload of human verifiers by providing them with evidence candidates so that their job turns from finding the evidence by themselves to verifying the correctness of the evidence acquired by machine. To address the missing-detail issue and the complicated semantic challenge, we propose a novel multi-modal neural-symbolic learning framework. Different from the existing neural-symbolic multi-modal models that convert all modalities to symbolic graphs, the proposed framework only parses the text to an abstract-meaning-representation graph\cite{banarescu2013abstract}. With the AMR graph and vision input, the proposed framework acquires interpretable predictions through the following steps: (1) First, a \textbf{neighbor-search-based algorithm} designed by us to extract a set of symbolic queries from the graph. (2) Then, the queries, together with the image, will be forwarded into a large pre-trained vision-language model (e.g., CLIP \cite{radford2021learning} and/or BLIP \cite{li2022blip}) to predict whether the image supports the queries or not, and (3) Finally, we develop an \textbf{evidence selection model} to rank the importance of each query and give the final prediction based on the query answers and their importance. And the query answers that have high importance scores and support the prediction will be output together as evidence.
Overall, our \textbf{contributions} are three-fold as follows:
\begin{itemize}
    \item We proposed an evidence-based paradigm for interpretable detection of cross-modal de-contextualization. It detects out-of-context misinformation based on extracted pieces of evidence, which eases the workload of human verifiers and makes the prediction of the detector more explainable and trustworthy.
    \item To address the drawbacks of existing neural-symbolic multi-modal models when confronted with misinformation detection, we designed a novel query-based framework that does not miss details in vision modality and incorporates the knowledge in large pre-trained models to capture complicated semantics.
   \item We conduct experiments to verify the accuracy of the model in mismatched text-image pairs detection and an empirical evaluation of the generated pieces of evidence. 
\end{itemize}

\begin{figure*}[!htb]
    \centering
    \includegraphics[width=6.5in]{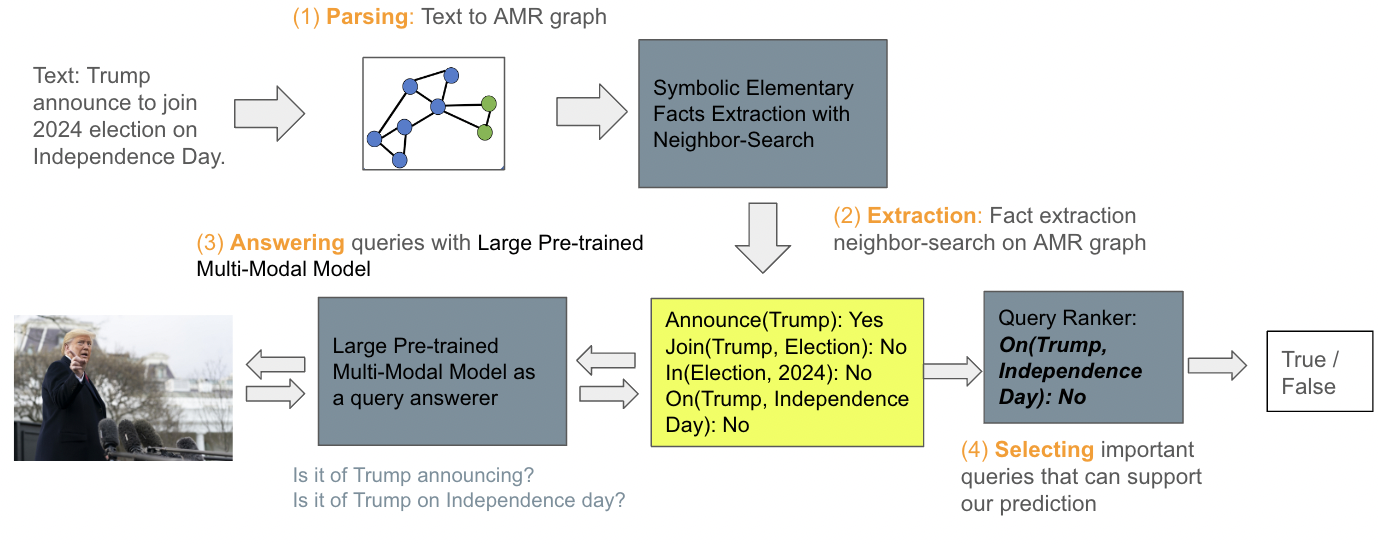}
    \caption{Overview of our proposed method. It first parses the text to AMR graphs with on-the-shelf tools. Then it extracts queries with a symbolic elementary fact extraction algorithm designed by us. After that, a large-pre-trained multi-modal model will determine whether the queries are supported by the vision input or not. Finally, a query ranker will select the important and reliable queries as the evidence to make the final judge.}
    \label{fig:overview}
\end{figure*}

\section{Related Works}
\textbf{Misinformation Detection:} Regular misinformation detection include single-modal detection and multi-modal detection. Single-Modal misinformation detection, which was a hot research topic, usually makes predictions based on some statistical features, such as traces of Photoshop or DeepFake in vision modality \cite{Trinh_2021_WACV}, and linguistic cues that appear more in fake text \cite{shu2017fake,qian2018neural}. And multi-modal misinformation generally aggregate the prediction, representations and/or intermediate results (e.g. feature maps) of singe-modal methods with fusion architectures, such as addition, concatenation and attention mechanisms to integrate the information extracted from different modalities and train a joint classifier \cite{khattar2019mvae,zhou2020similarity}. However, this paradigm brings the following problems. First, the misinformation creator can avoid being detected by using mismatched true images (videos) and changing their linguistic patterns. Second, such methods usually use deep neural networks to extract features which are lack of interpretability. Finally, the linguistic cues and diffusion patterns are not based on fact or logic. They are not sufficient for social media platforms and/or fact-checking websites to adopt action based on them. In contrast, our model provides interpretable prediction and extracted evidence from the perspective of factual contradictions. In this way, our model can detect misinformation even if both the text and vision inputs are drawn from the distributions of real news respectively.

\textbf{Neural-Symbolic Multi-Modal Learning:} Existing Neural-Symbolic Multi-Modal Learning methods are usually designed for Vision Question Answering \cite{yi2018neural,zhu2022generalization}. They first extract scene graphs \cite{johnson2015image} from vision modality and abstract-meaning-representation (AMR) graph \cite{banarescu2013abstract} from language modality, respectively. Then the model outputs interpretable answers by doing explicit reasoning on the extracted scene graph and text graph. However, such methods suffer from multiple drawbacks. First, training such models requires not only the label of the task (e.g., the answer of VQA and True/False for misinformation detection) but also scene graph annotation. Second, the existing scene graph parsers can only handle concepts in the training set and lack of zero-shot generalization ability. Third, for misinformation detection, the key points might be hidden in details that are often ignored by scene graph parsing tasks, e.g., background. Finally, but not lastly, for misinformation detection, we not only need basic concepts like actions and object labels but also complicated concepts like social relations and identity. In contrast, our proposed model only convert the text modality to semantic graphs to avoid detail missing and use a large pre-trained multi-modal models to capture the complicated semantics without requiring scene graph annotations.

\section{Proposed Method}
This paper aims to develop an interpretable cross-modal misinformation detection model that can jointly output prediction and the supporting ``evidence" based on the neural-symbolic method. Instead of using expensive scene graphs or implicit representations, we create a unifying and explicitly symbolic graph from both textual and visual information, reconciled by fuzzy logic. Each edge in the graph will be labeled with ``True" or ``False" based on the consistency between textual and visual information. A deep-learning-based ranker then scores each edge in the graph based on its reliability and importance to the full statement. The edges with high scores will be selected into the evidence sets. Finally, by counting the number of ``True" edges within the evidence sets, the model makes a prediction, and the edges that support the final prediction can be output as evidence. An overall figure is shown in Figure \ref{fig:overview}.  

\begin{algorithm}[!htb]

\caption{Synthetic Data Generation}
\begin{algorithmic}[1]
\REQUIRE AMR graph G = <V, E>
\ENSURE A set of elementary statements S
\STATE S=$\emptyset$
\FOR {n $\in$ V}
\IF{n.POS is Noun or Named\_Entity}
\IF{n is Time or Location}
\STATE S += <n,Spatial-Temporal>
\ELSE
\STATE S += <n,Object>
\ENDIF
\FOR {neighbor in n.neighbors}
\IF{neighbor.POS is Adjective}
\STATE S += <n, neighbor, {Attribute}>
\ENDIF
\ENDFOR

\ELSIF{n.POS is Verb}
\STATE L $\leftarrow$ NULL
\FOR {neighbor in n.neighbors}
\IF{neighbor.POS is Noun}
\STATE L+=neighbor
\ENDIF
\ENDFOR
\IF{L.length=1}
\STATE S += <L[0], n, {Activity}>
\ELSIF{L.length=2}
\STATE S += <L[0], n, L[1], {Relation}>
\ENDIF
\ELSIF{n.POS is Pronoun}
\STATE L $\leftarrow$ NULL
\FOR {neighbor in n.neighbors}
\IF{neighbor.POS is Noun}
\STATE L+=neighbor
\ENDIF
\ENDFOR
\IF{L.length=2}
\STATE S += <L[0], n, L[1], {Relation}>
\ENDIF
\ENDIF
\ENDFOR
\end{algorithmic}
\label{alg:extract}
\end{algorithm}

\subsection{Symbolic Graph and Query Generation}
In debating and discussion, human usually verify the correctness of and/or refute others' statement by splitting the whole statement to a set of elementary statements and then addressing them separately. Inspired by this human strategy, we propose to split the captions to elementary statements and identify whether the elementary statements are supported or contradicted with the vision modal information. We mainly consider 5 kinds of elementary statements that commonly exists in news: 
\begin{itemize}[leftmargin=*]
    \item Object Statement: The photo is about X (noun or entity in the caption);
    \item Spatial-Temporal Statement: The photo is taken in Y (a place or time);
    \item Activity Statement: The photo is about X (noun or entity in the caption) doing Y (verb);
    \item Relationship Statement: The photo is about X (noun or entity in the caption) doing Y (predicate, such as verb and pronouns) to Z (noun or named entity in the caption);
    \item Attribute Statement: In the photo, X (noun or entity in the caption) is Y (an attribute).
\end{itemize}

\begin{figure*}
    \centering
    \includegraphics[width=6.0in]{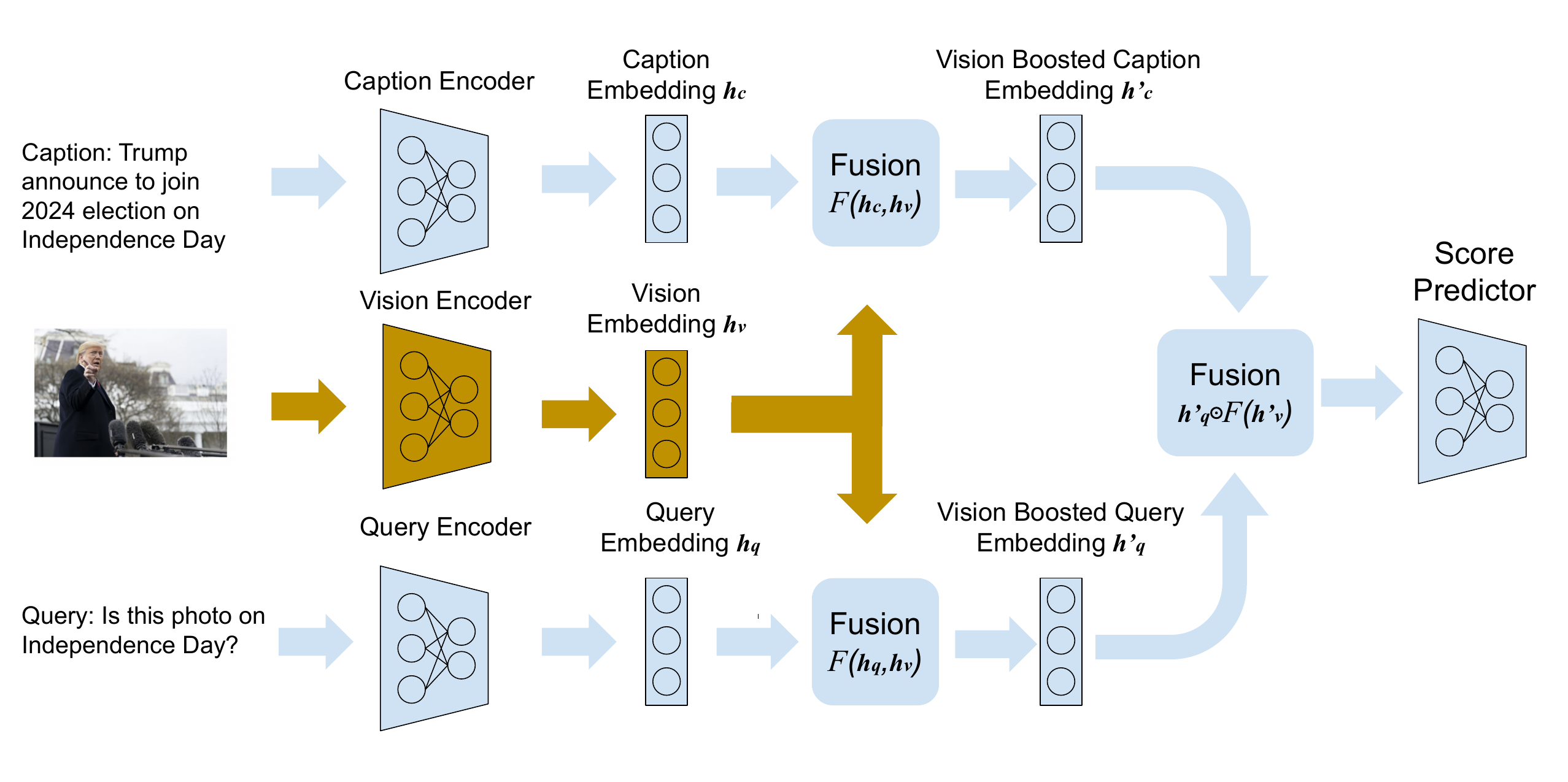}
    \caption{Pipeline of query ranker. The caption and query embeddings are respectively boosted by a vision embedding vector and then fused to acquire a final representation $h_f$. After that, $h_f$ is forwarded into a predictor to acquire the final evidence score.}
    \label{fig:query}
\end{figure*}

To extract the above elementary statements, we first conduct named entity recognition and abstract-meaning-representation (AMR) parsing on the caption to acquire an graph description, where each node is a word/phrase(e.g. noun, verb, pronoun and so on) or phrase (named entity) or semantic role (e.g. person, organization and so on) and each edge represent the grammar relationship between the connected node pair. On this graph, the 5 kinds of statements correspond to 5 kinds of node/paths:
\begin{itemize}[leftmargin=*]
    \item Object Statement: Noun or named entity;
    \item Spatial-Temporal Statement: Time or location;
    \item Activity Statement: the path of Subject-Verb (Subject can be noun or named entity);
    \item Relationship Statement: the path of Subject-Predicate-Object. Subject and Object can be nouns and/or named entities. Predicate can be verb or pronoun;
    \item Attribute Statement: the path of Adjective-Noun and Adjective-Named Entity.
\end{itemize}

The Object Statement and Spatial-Temporal Statement can be extracted by enumerating all nodes. And the other three kinds of paths can be extracted by doing specific neighbor search on all the nouns, named entities (including person, organization and so on), verbs, and pronouns. Detailed pseudo codes are shown in Appendix (Algorithm \ref{alg:extract}). After acquiring the elementary statements, we can convert them to the corresponding natural language queries (e.g. <X, Object> $\rightarrow$ Is the photo about X?) and forward them to a pre-trained large multi-modal model for answers. Based on the ratio of "Yes"/"No", we can predict the credibility of the information as well as output the evidence that supports our prediction.

\subsection{Query Ranker}
As discussed in the previous section, the answers for each query from the large vision-language model can be applied to compute the credibility scores of the text-image pair. However, such prediction suffers from relatively poor accuracy for three reasons. First, the large pre-trained multi-modal modal may give us wrong answers on some hard queries. Second, limited by the accuray of the AMR parsing and named entity recognition, some generated queries might be wrong. Finally, but not least, even for real news, not all elementary statements are contained and/or supported by the vision modality. For example, in sports news, the captions may simultaneously contain the final result of the game, but the image is just a fancy moment during the whole game. Due to the above reasons, we propose to design a query ranker to filter out those queries that are most likely to be supportive to the final prediction, i.e. when the query answer is "Yes", the caption is more likely to be consistent and when the query answer is "No", the caption is more likely to be inconsistent.

\begin{algorithm}[!htb]

\caption{Training of $f$ and $F_{enc}$}
\begin{algorithmic}[1]
\REQUIRE Training set $T=\{<v,c,q,A_q,l>\}$, where each sample contains vision input $v$, caption $c$, query $q$, query answer $A_q$, and label $l$ of the vision-caption pair. Triplet encoder $F_{enc}$, and classifier $f$.
\ENSURE Well-trained $F_{enc}$ and $f$
\FOR {$<v,c,q,A_q,l> \in T$}
\IF{$A_q=Yes$ and $l=True$}
\STATE $l'\leftarrow0$
\ELSIF{$A_q=Yes$ and $l=Fake$}
\STATE $l'\leftarrow1$
\ELSIF{$A_q=No$ and $l=True$}
\STATE $l'\leftarrow2$
\ELSE
\STATE $l'\leftarrow3$
\ENDIF
\STATE $loss=-\log(P(f(F_{enc}(v,c,q))=l'))$
\STATE Conduct back-propagation to compute gradient and run gradient descent.
\ENDFOR
\end{algorithmic}
\label{alg:train}
\end{algorithm}

The pipeline of the proposed model is shown in Figure \ref{fig:query}. Given a triplet containing an image, a caption and a query, three encoders first represent them respectively as three embedding vectors $h_q$, $h_c$ and $h_i$, which share the same dimension to enable flexible choices on the embedding fusion in the next step. After that we boost the caption embedding and query embedding respectively with the image embedding as follows:

\begin{equation}
    h_{qb} = F_{agg}(h_q,h_v), h_{cb} = F_{agg}(h_c,h_v)
\end{equation}
where $h_q$, $h_c$ and $h_v$ are the embeddings of query, caption and vision input respectively, $h_qb$, $h_cb$ are the boosted query and caption embedding respectively, and $F_{agg}$ is an aggregation function that fuse two embeddings. Based on empirical result, we select pair-wise multiply as $F_{agg}$. After that, we fuse the two vision-boosted embeddings as follows:
\begin{equation}
    h_{f} = h_{qb}\odot F_M(h_{cb})
\end{equation}
where $\odot$ is pair-wise multiply of two vectors, and $F_M$ is a trainable mapping function that maps the $h_{cb}$ into the same space of $h_{qb}$. Here, we apply a multi-layer perceptron (MLP) as $F_M$. With $h_f$ which aggregate information from caption, query and vision input, we can predict the supporting probability $P_S(q)$ of query $q$. A query is considered as a supportive query if its answer is consistent with the credibility of the news, i.e. ``query answer is Yes" $\rightarrow$ ``True News" and ``query answer is No" $\rightarrow$ ``Fake News"). From this definition, we can rewrite the probability $P_S(q)$ that the query $q$ is supportive as:
\begin{equation}
    P_S(q) = P(True, A_q=Y)+P(Fake,A_q=N)
\label{eq:add}
\end{equation}

\begin{table*}[!htb]
    \centering

\begin{tabular}{|l|c|c|c|c|}
\hline
\textbf{Method} & \textbf{Accuracy}$\uparrow$ & \textbf{AUC of ROC}$\uparrow$& \textbf{FAR}$\downarrow$ & \textbf{FRR}$\downarrow$\\
\hline
ARCNN-MLPs & 53.9 & 56.2 & 39.1 & 52.9 \\
SAFE & 50.7 & -- & 58.4 & 60.1 \\
\hline
VisualBERT & 54.8 & -- & 54.9 & 35.4 \\
FaceNet + BERT & 59.6 & 63.7 & 40.3 & 40.6 \\
CLIP & 62.6 & 67.2 & 37.3 & 37.3 \\
VINVL & 65.4 & 71.9 & 34.2 & 34.2 \\
DT-Transformer & 65.7 & - & \textbf{26.3}& 42.4 \\
SSDL & 65.6 & - & 31.6& 37.2 \\
\hline
BLIP-2 (single best prompt) & 61.6 & -- & 46.7 & \textbf{30.0} \\
BLIP-2 (prompt ensemble) & 62.4 & 65.2 & 43.3 & 32.0 \\
\hline
Ours (w/o Query Ranker) & 62.8 & 66.9 & 42.8 & 33.5 \\
Ours & \textbf{68.2} & \textbf{73.0} & 29.5 & 34.6 \\
\hline
\end{tabular}
\caption{Comparison between the baselines and our proposed model. 
}
\label{tab:comparison}
\end{table*}
where $P(True, A_q=Y)$ is the probability that the news is true and the query answer is ``Yes", and $P(Fake, A_q=N)$ is the probability that the news is fake and the query answer is ``No". Inspired by this formula, we propose to train a 4-class classifier $f$ ($0\rightarrow <True, A_q=Y>$, $1\rightarrow <Fake, A_q=N>$, $2\rightarrow <True, A_q=N$>, $3\rightarrow <Fake, A_q=Y>$) that takes $h_f$ as input, and then use Equation \ref{eq:add} to predict the supportive probability as:
\begin{equation}
    P_S(q) = P(f(h_f)=0)+P(f(h_f)=1)
\end{equation}
The detailed training algorithms of the query ranker are in Algorithm \ref{alg:train}

\section{Experiments}

\subsection{Dataset}
We evaluate the performances of our models and baselines on an automated out-of-context misinformation detection benchmark NewsCLIPpings \cite{luo2021newsclippings}. NewsCLIPpings is built up based on VisualNews \cite{liu-etal-2021-visual}, an image-captioning dataset that collects image-text pairs from four news agencies: BBC; USA Today; The Guardian; and The Washington Post. NewsCLIPpings first constructs an image pool by extracting all news images from VisualNews and then generates automated out-of-context misinformation by using CLIP to retrieve semantically similar images from the pool for each text. Following, we report the results on the Merged-Balance version of this dataset. The official training/validation/testing ratio of this version is 10:1:1.

\subsection{Baselines}
We compared our methods with two kinds of baselines in this task:
\begin{itemize}
    \item Baselines trained from scratch: MLPs and SAFE \cite{zhou2020similarity} are two multi-modal misinformation detectors that are proposed before the era of large pre-trained multi-modal models. Therefore, they are all designed to be trained from scratch. We report their performances to present how poor multi-modal misinformation detectors will perform facing automated out-of-context misinformation when the knowledge from pre-training model is not incorporated.
    \item Baselines incorporated with large pre-trained multi-modal models: VisualBERT \cite{li2019visualbert}, CLIP \cite{luo2021newsclippings}, VINVL \cite{huang2022text}, SpotFake (FaceNet+BERT) \cite{singhal2019spotfake}, DT-Transformer \cite{papadopoulos2023synthetic}, and SSDL \cite{mu2023self} are pre-trained models fine-tuned on the NewsCLIPpings dataset. For VisualBERT, CLIP and VINVL, we followed the results reproduced by \citeauthor{huang2022text}. For SpotFake, we applied FaceNet for the vision input since we found that on many mismatched samples can be recognized through the identity of the people in the picture. The hyper-parameters of fine-tuning is set the same as the original paper of BERT In addition, we also report the performances of BLIP-2 with prompt engineering. We did not fine-tuned BLIP-2 on this dataset due to its large scale. For the prompt of BLIP-2, we applied two strategies: (1) the result with the best prompt we found, and (2) ensemble the results from different prompts.
\end{itemize}

\subsection{Quantative Results}
We report the comparison of our proposed method and baselines in Table \ref{tab:comparison}. From the table, we can observe the following phenomena. First, all the baselines trained from scratch perform poorly. Their accuracies are all close to 50\%, which is the expectation of a random baseline on the binary classification task. This phenomenon reflects how deceptive the automated out-of-context misinformation generated by large pre-trained multi-modal models can be to typical deep learning detectors. This is because the automated out-of-context can usually guarantee the basic semantic consistency between images and texts. The inconsistency usually appears in some details, such as location, seasons and identities of the appeared people in the photo. Without the knowledge from large pre-training models, such detailed inconsistencies are hard to detect. Second, our the proposed method outperform all baselines on almost all metrics, indicating the effectiveness of the propose model. It is also noticeable that our model's variant without Query Ranker can achieve a performance that is approximately same as fine-tuned CLIP, but still get substantially outperformed by VINVL. This reflect the limitations of AMR parsers and BLIP-2, which can respectively introduce errors into query extraction and query answering, and indicate the importance of query ranker.

\subsection{Evaluation on Interpretability}
In this sub-scetion, we aim at measuring the interpretability of our models by evaluating the quality of the ``evidences" generated by it. To this end, we annotate a fraction test set with the their ground-truth evidences, i.e the logical contradiction betweetn the text and the image. More specifically, we ask annotators to first learn our definition to the 5 kinds of cross-modal factual mistakes and then label the data samples based on the template in Sec. 3.1. In this process, we require the annotators to use the original words in the text to fill in the X, Y, Z in the templates for convenient evaluation.

In this experiment, since the baselines are not interpretable models, we apply an on-the-shelf deep learning model interpretor to identify the word phrases that contributes most to their prediction. We report the HIT@top10 score. We consider an evidence raised by the model as a "hit" evidence if at least one of its word appears in the X, Y, or Z. The results are shown in Table \ref{tab:inter}. As we can see, by applying on-the-shelf deep learning model interpretors, the baselines can raise some evidences with reasonable quality (i.e. substantially better than random). However, compared to our model which make the prediction in an inherently explainable way, their performances are significantly lower. And compared to the variant without query ranker, our full model shows better score, indicating that the query ranker help the model to refine the quality of the raised evidences.

\begin{table}[!htb]
    \centering
    \begin{tabular}{c|c}
    \hline
        Model & HIT@10 \\
        \hline
        Random Baseline & 12.4\\
        CLIP & 19.8\\
        FaceNet + BERT & 22.2\\
        Ours w/o Query Ranker &35.3\\
        Ours & 38.1\\
        \hline
    \end{tabular}
    \caption{Result of the HIT@10 for interpretability evaluation.}
    \label{tab:inter}
\end{table}

\section{Limitations}
The proposed model is still a small leap toward interpretable, and furthermore procedural-justice, misinformation detection. It only considers factual inconsistency between language modality and vision modality, which only works for misinformation containing factual errors, such as fake or distorted news. However, it may not work when addressing other kinds of disinformation, such as hate speech meme and political propaganda. Besides, for now the proposed model can only provide the coarse language-originated evidences, i.e. which query is not consistent with the image. However, in practice, human verifiers may also hope to acquire a more detailed evidences, such as which region in the image leads to the inconsistency, and/or references that can help them document the clarification.

\section{Conclusion}
In this paper, we proposed an interpretable cross-modal de-contextualization misinformation detector. It applies a neural-symbolic model to extract factul queries and make prediction based on the query answers from large pre-trained multi-modal model. Compared to existing models, the proposed model not only provides predictions on the credibility of the news, but also gives supporting evidences. Our experiment results indicate that the proposed model not only achieves a competitive performance, but also provide better interpretability.

For future works, an important direction to explore is how to organize the extracted evidences as convincing clarification text, which can further increase the efficiency and immediacy. Moreover, as we discussed in the limitation section, the current extracted evidences only help us localize the factual inconsistency in the language modality. However, the image regions that leads to inconsistency are also important. Therefore, in the furture, we will consider how to detect the inconsistent pair of an image region and a text phrase.

\bibliography{emnlp2022}
\bibliographystyle{acl_natbib}




\end{document}